\documentclass[final]{l4dc2026}
\usepackage{times}
\usepackage{nicefrac}

\usepackage{graphicx} %
\usepackage{mwe}
\usepackage{subcaption}
\usepackage{multirow}
\usepackage{multicol}
\usepackage{nicematrix}
\usepackage{booktabs}
\usepackage{float}
\usepackage{tikz}
\usepackage{mathtools}
\usepackage{amsmath}
\usepackage{amssymb}
\usepackage{bm}
\usepackage{xcolor}
\usepackage{todonotes}
\usepackage{cleveref}
\usepackage{amsmath}
\usepackage{enumitem}

\DeclareMathOperator*{\argmin}{argmin\,}

\newcommand{\R}{\mathbb R}
\newcommand{\E}{\mathbb E}

\newcommand{\Nc}{\mathcal N}

\title[Probabilistic surrogate modeling techniques for partially-observed systems]{Efficient probabilistic surrogate modeling techniques for partially-observed large-scale dynamical systems}

\author{%
 \Name{Hans Harder} \Email{hans.harder@uni-paderborn.de}\\
 \addr University of Paderborn \& Lamarr Institute for Machine Learning and Artificial Intelligence
 \AND
 \Name{Abhijeet Vishwasrao} \Email{abvish@umich.edu}\\
 \addr University of Michigan%
 \AND
 \Name{Luca Guastoni} \Email{luca.guastoni@tum.de}\\
 \addr Technical University of Munich%
 \AND
 \Name{Ricardo Vinuesa} \Email{rvinuesa@umich.edu}\\
 \addr University of Michigan%
 \AND
 \Name{Sebastian Peitz} \Email{sebastian.peitz@tu-dortmund.de}\\
 \addr Technical University of Dortmund \& Lamarr Institute for Machine Learning and Artificial Intelligence%
}

\begin{document}

\maketitle

\begin{abstract}
    This paper is concerned with probabilistic techniques for forecasting dynamical systems described by partial differential equations (such as, for example, the Navier--Stokes equations). In particular, it is investigating and comparing various extensions to the flow matching paradigm that reduce the number of sampling steps. %
    In this regard, it compares direct distillation, progressive distillation, adversarial diffusion distillation, Wasserstein GANs and rectified flows. Moreover, experiments are conducted on a set of challenging systems. %
    In particular, we also address the challenge of directly predicting 2D slices of large-scale 3D simulations, paving the way for efficient inflow generation for solvers.
\end{abstract}


\maketitle

\section{Introduction}

Partial differential equations (PDEs) are used to describe physical phenomena such as heat transfer, fluid flows, or wave propagation in different media. Solving these equations requires high-resolution spatial discretization and accurate time stepping schemes, leading to high computational cost. This hinders the availability of fast forecasting models based on classical numerical approaches. To address these shortcomings, approximate solutions with data-driven surrogate models have been developed, for example in the case of meteorology \citep{lamLearningSkillfulMediumrange2023,priceProbabilisticWeatherForecasting2025}, climate \citep{watt-meyerACEFastSkillful2023} or ocean dynamics \citep{chattopadhyayOceanNetPrincipledNeural2024}.
These models are typically trained in a supervised fashion to match the output of the numerical solver or real data. They are defined on the same or a downscaled version of the solver's spatial mesh~\citep[even though there are also mesh-agnostic methods, \textit{e.g.},][]{liNeuralOperatorGraph2020, liFourierNeuralOperator2021a} and usually predict larger time steps.

Learning approximate solutions becomes a necessity when the target dynamical system is only partially observable or when the mathematical description of the problem does not perfectly capture the underlying physical system (\textit{e.g.}, when a model for weather forecasting ignores local topology). While the latter problem can be addressed by conditioning on real data, the former appears more challenging. One way to approach this issue is to include information about the dynamics from past events using, for instance, recurrent network architectures \citep[\textit{e.g.,}][] {PhysRevFluids.4.054603,vlachasDatadrivenForecastingHighdimensional2018,frommeSurrogateModeling3D2025}, motivated in part by Taken's embedding theorem \citep{takensDetecting1980}. But these models are typically hard to train, require many training samples and access to complete trajectories.

Alternatively, it is possible to lean into the lack of information by learning a probabilistic model instead: With the recent advent of denoising diffusion \citep{sohl-dicksteinDeepUnsupervisedLearning2015,hoDenoisingDiffusionProbabilistic2020} and flow matching \citep{lipmanFlowMatchingGenerative2023}, high-quality generative models have become widely accessible and easy to train. However, using these models comes with a caveat: To generate a new sample, one needs to solve an ordinary differential equation (ODE), requiring multiple evaluations of the trained model. This is a relatively general issue, so there have been efforts to improve the sampling speed. These approaches, further detailed in \Cref{sec:improving-sampling-efficiency}, range from training a new model to learn the ODE's flow via ``distillation'' to methods that straighten the underlying velocity field and, even more recently, adversarial approaches that introduce discriminator networks into training.

Our main goal is to demonstrate that flow matching together with these extensions poses a serious alternative to deterministic surrogate modeling techniques.
Insofar, we want to show that
\begin{itemize}[itemsep=0em]
    \item these methods generate trajectories that correctly approximate the real solutions, are temporally coherent, and closely match the reference as the setting becomes more deterministic,
    \item further distillation/rectification speeds up the sampling process, allowing single-step sampling in some cases, while being superior in physical accuracy to GANs,
    \item these methods %
    open up entirely new possibilities for prediction or re-initialization of expensive solvers. We demonstrate this with experiments on the compressible Navier-Stokes (NS) equations and the Rayleigh-Taylor instability (RTI) in two and three dimensions.
\end{itemize}
The code for this paper is publicly available under \href{https://github.com/graps1/flow-matching-for-time-series}{github.com/graps1/flow-matching-for-time-series}.

\subsection{Related work and discussion}\label{sec:related-work}

Denoising diffusion, score matching and flow matching progressively denoise samples from an initial Gaussian noise distribution. These methods were found to be largely equivalent~\citep{holderrieth2025generatormatchinggenerativemodeling}. %
Diffusion probabilistic models were developed by \cite{sohl-dicksteinDeepUnsupervisedLearning2015}, which were popularized and found to be equivalent to score matching by \cite{songGenerativeModelingEstimating2019, hoDenoisingDiffusionProbabilistic2020, dhariwalDiffusionModelsBeat2021, nicholImprovedDenoisingDiffusion2021}. 
Denoising diffusion ``implicit'' models \citep{songDenoisingDiffusionImplicit2021} made the first step towards a deterministic sampling process (except for the initial noise sample) by formulating the solution as an ordinary instead of a stochastic differential equation. In the same spirit, flow matching \citep{lipmanFlowMatchingGenerative2023,lipmanFlowMatchingGuide2024} popularized the ``Gaussian optimal transport path'', an advantageous coupling of noise and target distribution. %

In these years, a lot of progress has been made in speeding up the sampling process: First the transition from SDEs to ODEs, then the popularization of better transport couplings. However, there is still room for improvement when comparing the sampling speed to that of GANs. Insofar, the recent literature provides us with a few approaches: Distillation-based methods such as consistency distillation \citep{songConsistencyModels2023,songImprovedTechniquesTraining2024} or progressive distillation \citep{salimansProgressiveDistillationFast2022} learn to predict the ODE's solution from the initial sample. One can also employ adversarial methods by including a discriminator loss in the distillation \citep{linSDXLLightningProgressiveAdversarial2024, sauerAdversarialDiffusionDistillation2023} or diffusion \citep{xiaoTacklingGenerativeLearning2022} procedure. Finally, rectified flows \citep{liuFlowStraightFast2023,liuInstaFlowOneStep2024} ``straighten'' the underlying flow, allowing larger step sizes when solving the ODE. 

Diffusion models have shown great potential for the surrogate modeling of dynamical systems, including turbulent flows~\citep{Guastoni2025,vishwasraoDiffSPORTDiffusionbasedSensor2025}. Most similar to our work are the papers by \cite{liGenerativeLatentNeural2025, luoDiffFluidPlainDiffusion2024, priceProbabilisticWeatherForecasting2025, oommenLearningTurbulentFlows2025, kohlBenchmarkingAutoregressiveConditional2024} and \cite{shysheyaConditionalDiffusionModels2024}. They all describe methods for forecasting systems autoregressively using conditional diffusion models, either in latent-spaces \citep{liGenerativeLatentNeural2025}, in the fully-observed setting \cite{luoDiffFluidPlainDiffusion2024,oommenLearningTurbulentFlows2025,kohlBenchmarkingAutoregressiveConditional2024} or for partially observed/real-world data \cite{shysheyaConditionalDiffusionModels2024,priceProbabilisticWeatherForecasting2025}. Further work includes the paper by \cite{yangDenoisingDiffusionModel2023}, which predicts the system's solution given only the initial condition, and \cite{cachayDYffusionDynamicsinformedDiffusion2023}, which couples the denoising and forecasting processes. The paper of \cite{el-gazzarProbabilisticForecastingAutoregressive2025} is dealing with forecasting ordinary differential equations. It is also worth noting that one can also use a deterministic latent-space model and instead reconstruct the full state using a diffusion model \citep{gaoGenerativeLearningForecasting2024}. 

In contrast, our focus is on accelerating the sampling of flow matching-based models.

\begin{figure}
    \centering
    \includegraphics[width=\linewidth]{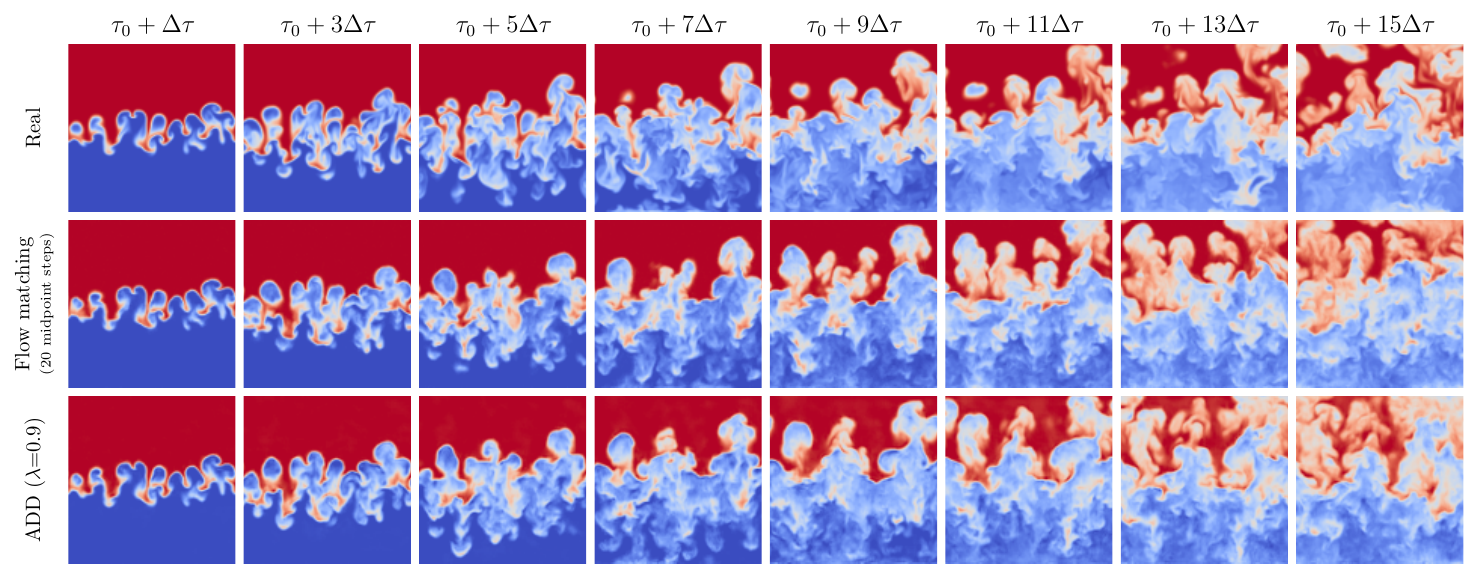}
    \caption{Autoregressively generated trajectories on the ``sliced'' Rayleigh-Taylor instability dataset, starting with the same initial condition at (simulation) time $\tau_0$.}
    \label{fig:showcase-rti}
\end{figure}

\section{Flow matching}

Flow matching (FM, \cite{lipmanFlowMatchingGenerative2023, lipmanFlowMatchingGuide2024}) defines a continuous transformation from samples of a Gaussian distribution $\bm x_0 \sim p_0 = \Nc(0,I)$ at time $t = 0$ to samples of some target distribution $\bm x_1 \sim p_1$ at time $t = 1$. Both $\bm x_0$ and $\bm x_1$ take values in $\R^n$. The starting point is the definition of an interpolating marginal distribution $p_t(x_t)$, which is classically defined as the the distribution of
\begin{equation}
    \bm x_t = t \bm x_1 + (1-t) \bm x_0 =: \psi_t(\bm x_0 | \bm x_1),
    \label{eq:gaussian-optimal-transport-path}
\end{equation}
where $\bm x_0$ and $\bm x_1$ are samples from $p_0$ and $p_1$.
Equation \eqref{eq:gaussian-optimal-transport-path} defines the \emph{Gaussian optimal transport path}, which is simply a linear interpolation between source and target samples.
The goal is to learn the velocity field $\dot x_t = u_t(x_t)$ that pushes $p_t(x_t)$ forward in time: If a particle tracks the velocity field $u_t(\bm x_t)$ such that the initial condition is sampled from $\bm x_0 \sim p_0$, then $\bm x_t$ tracks $p_t$ in distribution until it finally reaches the target distribution at $t = 1$. Remarkably, if $v_t^\theta(x_t)$ ($v^\theta : [0,1] \times  \R^n \rightarrow \R^n$) is a neural network parametrized by $\theta$, one can fit it to $u_t(x_t)$ by solving the optimization problem
\begin{equation}
    \min_\theta \mathbb E_{p_1(\bm x_1), p_0(\bm x_0), U(\bm t; 0,1)} \lVert \underbrace{v_{\bm t}^\theta(\bm x_{\bm t}) - \dot {\bm x}_{\bm t}}_{\mathclap{=v_{\bm t}^\theta(\bm t \bm x_1 + (1-\bm t) \bm x_0) - (\bm x_1 - \bm x_0)}} \rVert^2, \quad\text{where}\quad \bm x_{\bm t} =\psi_{\bm t}(\bm x_0 | \bm x_1).
    \label{eq:flow-matching-problem}
\end{equation}
Approximate solutions can be obtained by discretizing in time, for instance by using the explicit Euler method.
After training, the inference process consists in sampling from the initial distribution $\bm x_0 \sim p_0$ and then solving for $\bm x_1 = \phi^\theta_1(\bm x_0)$, where $\phi_t^\theta(x_0)$ is the solution of the learned ODE at time $t$, i.e., $\phi_t^\theta(x_0)=x_t$ if $\dot x_t = v_t^\theta(x_t)$.

When dealing with a \emph{conditional target distribution} $\bm x_1  \sim p_1(\cdot | y)$, the training procedure changes minimally. One adds $y$ as an additional input to the velocity network, which becomes $v_t^\theta(x_t | y)$, and one assumes a joint sampling distribution $(\bm x_1, \bm y) \sim p_1(\cdot, \cdot)$. The problem one solves is then
\begin{equation}
    \min_\theta \mathbb E_{p_0(\bm x_0), p_1(\bm x_1, \bm y), U(\bm t; 0, 1)} \lVert v_{\bm t}^\theta(\bm x_{\bm t} | \bm y) - \dot {\bm x}_{\bm t} \rVert^2, \quad\text{where}\quad \bm x_{\bm t} =\psi_{\bm t}(\bm x_0 | \bm x_1). \label{eq:flow-matching-conditional}
\end{equation}
As for the unconditional case, we denote solutions by  $\phi^\theta_t(x_0|y) = x_t$ if $x_t$ tracks the conditional flow matching ODE, i.e., $\dot x_t = v_t^\theta(x_t|y)$.

\subsection{Flow matching for dynamical systems} 

We model transitions between partial observations as a Markov chain
\begin{equation}
    \bm y_{k+1} \sim p_1(\cdot | \bm y_k), \quad k = 0, 1,2, \dots
    \label{eq:markov-chain}
\end{equation}
with $\bm y_0$ stemming from some initial distribution, dependent on the problem at hand. 
The density $p_1(\cdot | y)$ for a given $y$ is the conditional target distribution of a flow matching model. In other words, we are learning a conditional flow matching velocity model $v_t^\theta(x_t|y)$ such that transitioning via
\begin{equation}
    \bm y_{k+1} = \phi^\theta_1(\bm x_{0}^k |\bm y_k), \quad \bm x_{0}^k \sim p_0 =\Nc(0,I)
\end{equation}
yields approximately the same distribution as \eqref{eq:markov-chain}.

\subsection{Deterministic models \& the initial flow matching direction}\label{sec:deterministic-models}

Before discussing any extensions to flow matching, we want to argue why it is a good fit for our applications. In particular, we want to show that it becomes more efficient as the underlying distribution becomes more concentrated, i.e., as transitions become more deterministic, in the sense that a single Euler step resembles the output of a deterministic surrogate model.

\begin{figure}
    \centering
    \includegraphics[width=\linewidth]{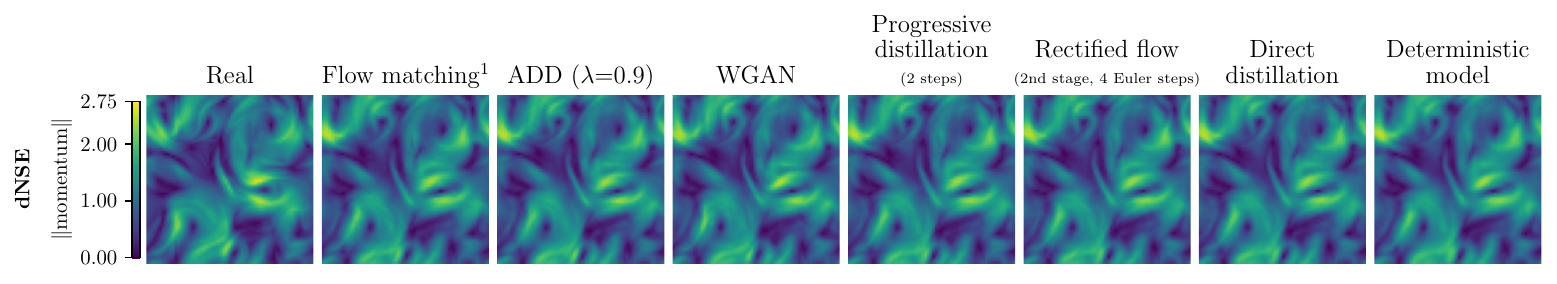}
    \includegraphics[width=\linewidth]{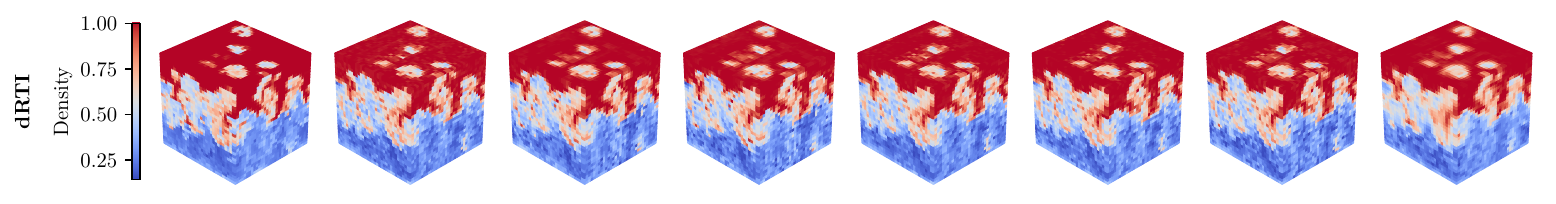}
    \includegraphics[width=\linewidth]{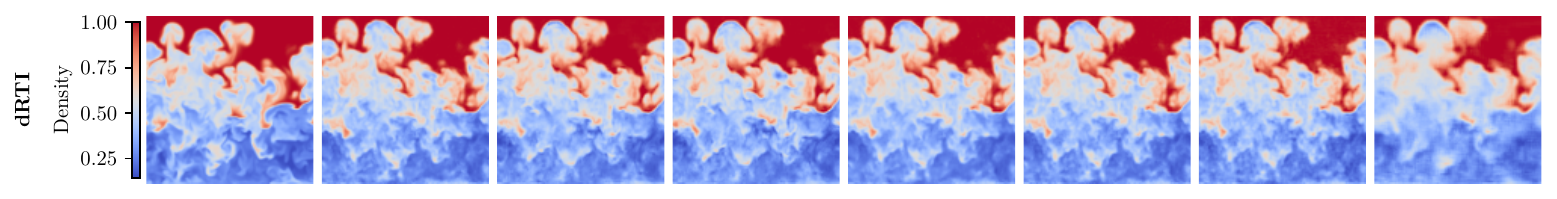}
    \caption{Results after a single prediction step. $^1$FM solved with 5 midpoint steps on the \textbf{dNSE} and \textbf{dRTI} datasets, and with 20 midpoint steps on the \textbf{sRTI} dataset.}
    \label{fig:single-prediction}
\end{figure}

Suppose therefore we were to train a deterministic predictor $w^\theta : \R^n \rightarrow \R^n$ by minimizing a least-squares problem
\begin{equation}
    \E_{p_1(\bm y_{k+1}, \bm y_k)} \lVert w^\theta(\bm y_k) - \bm y_{k+1} \rVert^2.
    \label{eq:deterministic-optimization-problem}
\end{equation}
It turns out that \eqref{eq:deterministic-optimization-problem} is up to an additive constant identical to the objective
\begin{equation}
    \E_{p_1(\bm y_k)} \lVert w^\theta(\bm y_k) - \E_{p_1(\bm y_{k+1}|\bm y_k)}[ \bm y_{k+1} ] \rVert^2,
    \label{eq:deterministic-optimization-problem-equivalent}
\end{equation}
see Remark \ref{rem:proof-of-equivalent-optimization-problems} in the appendix for further discussion.
In other words: A deterministic model that is trained by minimizing \eqref{eq:deterministic-optimization-problem} is implicitly fitted to the expected value of $\bm y_{k+1}$ (given $\bm y_k$). Since the expected value is a convex combination of all possible $\bm y_{k+1}$, it tends to be relatively smooth, a property that we also observe in our experiments (e.g., \Cref{fig:single-prediction}).
The target \eqref{eq:deterministic-optimization-problem-equivalent} can now be connected to the initial direction provided by a flow matching model. Starting with the flow matching objective \eqref{eq:flow-matching-conditional}, let us fix $\bm x_1 := \bm y_{k+1}$, $\bm y := \bm y_k$ and consider the objective at time $\bm t = 0$,
\begin{equation}
    \E_{p_1(\bm y_{k+1}, \bm y_k), p_0(\bm x_0)}\lVert v_0^\theta(\bm x_0|\bm y_k) - (\bm y_{k+1} - \bm x_0) \rVert^2,
    \label{eq:initial-direction}
\end{equation}
which has the same structure as \eqref{eq:deterministic-optimization-problem}. Similarly, it is up to a constant identical to the objective
\begin{equation}
    \E_{p_1(\bm y_k), p_0(\bm x_0)}\lVert v_0^\theta(\bm x_0|\bm y_k) - (\E_{p_1(\bm y_{k+1}|\bm y_k)}[\bm y_{k+1}] - \bm x_0)\rVert^2.
    \label{eq:initial-direction-equivalent}
\end{equation}
This loss is zero when
\begin{equation}
    x_0 + v_0^\theta(x_0|y_k) = \E_{p_1(\bm y_{k+1}|y_k)}[ \bm y_{k+1}] \quad \text{for almost all } (x_0, y_k).
    \label{eq:initial-direction-explicit-euler}
\end{equation}
Crucially, the expression on the left-hand side of \eqref{eq:initial-direction-explicit-euler} is also computed when solving the flow matching ODE with a single explicit Euler step. This shows that a well-trained flow matching model has the capacity to ``mimic'' a deterministic model, making it highly efficient for fully observed systems. Moreover, it shows that flow matching provides a good baseline for further extensions, which are discussed in the next section.

\subsection{Improving the sampling efficiency} \label{sec:improving-sampling-efficiency}

For brevity, we will only discuss methods for the unconditional case, since their extension to conditional target distributions is straightforward.

\paragraph{Direct distillation} learns a new model $G^\xi : \R^n \rightarrow \R^n$, parametrized by $\xi$, that predicts the solution of the flow matching ODE from the inital random sample. By direct distillation we understand methods that directly fit $G^\xi$ to $\phi^\theta_1$ in, for example, a least-squares sense:
\begin{equation}
    \min_\xi \E_{p_0(\bm x_0)} \lVert G^\xi(\bm x_0) - \phi_1^\theta(\bm x_0) \rVert^2. \label{eq:direct-distillation}
\end{equation}
A reasonable way to define $G^\xi$ for a flow matching model is to take $G^\xi(\bm x_0)= \bm x_0 + v^\xi_0(\bm x_0)$, where $\xi$ is initialized as a copy of $\theta$, resembling an explicit Euler step.
Optimizing \eqref{eq:direct-distillation} requires solving the flow matching ODE for every training sample, which can be challenging when many discretization steps are necessary. 

\paragraph{Progressive distillation~\citep{salimansProgressiveDistillationFast2022}} addresses this problem by learning the ODE's solution at increasing time steps. It keeps a ``student'' and an ``expert" model, where the former turns into the latter after each training run. The student model learns to predict two consecutive steps of the expert model, say the student learns to make a step of size $\delta$ whereas the expert makes two steps of size $\delta/2$. The step sizes are exponentially increasing ($\delta = \dots, \textstyle \frac 1 8, \frac 1 4, \frac 1 2, 1$) until they reach a final value of $1$. At that point, the student approximately solves the ODE in a single step. Though originally formulated in the DDIM (\cite{songDenoisingDiffusionImplicit2021}) context, one can adapt this idea to flow matching by defining the following iteration: Start with some $N > 0$ that is a power of $2$, e.g. $N = 32$, corresponding to an initial step size of $1/N$. As $N$ becomes larger, the Euler iteration
$$
    G^{\xi_N}_t(x_t) := x_t + \textstyle\frac 1 N v^{\xi_N}_t(x_t) 
$$
with $\xi_N = \theta$ tracks the flow more and more accurately. Starting with $m = N$, one can then solve the optimization problem
\begin{equation}
    \xi_{m/2} := \argmin_{\xi} \E_{p_{\bm t}(\bm x_{\bm t}), U(\bm t; \{1/k,2/k,\dots,1-2/k\})} \lVert G^\xi_{\bm t}(\bm x_{\bm t}) - G^{\xi_{k}}_{\bm t}(G^{\xi_{k}}_{\bm t}(\bm x_{\bm t})) \rVert^2,
\end{equation}
halving $m$ with each training run. Upon reaching $m = 1$, the ODE is approximately solved by $G_0^{\xi_1}$. In contrast to direct distillation, this procedure is more efficient, but also biased. With each training run, the model is fitted to targets produced by another model, which can lead to growing inaccuracy.

\paragraph{Adversarial diffusion distillation \citep[ADD,][] {sauerAdversarialDiffusionDistillation2023,linSDXLLightningProgressiveAdversarial2024}} is a distillation method that includes a discriminator loss, trained GAN-style by identifying the difference between fake (produced by the distilled model) and real samples. The discriminator $D^\zeta : \R^n \rightarrow \R$ approximates the Wasserstein-1 distance between the fake and the true distribution, substituting the hard Lipschitz constraint by a soft gradient penalty.
We use a similar setup as implemented by geometric GANs \citep{limGeometricGAN2017,sauerAdversarialDiffusionDistillation2023}, in which the discriminator aims to minimize
\begin{equation}
    \E_{p_0(\bm x_0)} [ \max \{ 0, 1 - D^\zeta(G^\xi(\bm x_0)) \}] + \E_{p_1(\bm x_1)}[\max \{ 0, 1 + D^\zeta(\bm x_1) \} + \gamma \lVert \nabla D^\zeta(\bm x_1) \rVert^2],
\end{equation}
where $\gamma \geq 0$ is the weight of the gradient penalty. The generator minimizes
\begin{equation}
    \lambda\E_{p_0(\bm x_0)} \rVert G^\xi(\bm x_0) - \phi^\theta_1(\bm x_0) \rVert^2 + (1-\lambda) \E_{p_0(\bm x_0)}[ D^\zeta(G^\xi(\bm x_0))],
\end{equation}
where $\lambda \in [0,1]$ weights the distillation loss. For the case $\lambda = 0$, this trains a geometric Wasserstein GAN (WGAN), whereas $\lambda = 1$ corresponds to direct distillation. In contrast to \cite{sauerAdversarialDiffusionDistillation2023}, the generator is only trained on samples from the initial noise distribution ($t = 0$), not on intermediate ones ($t > 0$).
Both generator and discriminator are based on a pre-trained flow matching model. In particular, we define them as
\begin{align}
    G^\xi(x_0) = x_0 + v_0^\xi(x_0) \quad\text{and}\quad D^\zeta(x_0) = \textstyle\sum_{i=1}^n \big( v_0^\zeta(x_0) \big)_i,
\end{align}
with the parameters $\xi$ and $\zeta$ initialized as copies of $\theta$.
Notably, the discriminator learns to distinguish between samples from the \emph{true} data distribution and fake samples, which means that it is not limited by the output quality of the flow matching model.
When training an ADD model, it is important that the discriminator learns faster than the generator, as it is part of the optimization problem's ``inner loop''.  One typically achieves this by choosing a larger learning rate compared to the generator, or alternatively by iterating 5-10 discriminator steps for each generator update.

\paragraph{Rectified flows~\citep{liuFlowStraightFast2023,liuInstaFlowOneStep2024}} 
change the ``coupling'' of the variables $\bm x_0$ and $\bm x_1$ from an independent joint distribution into one that has lower transport cost by retraining (rectifying) the flow matching model. %
After training a set of parameters $\theta_0$ with the classic flow matching formulation \eqref{eq:flow-matching-problem}, one iterates the following optimization problem:
\begin{equation}
    \theta_{m+1} := \argmin_\theta \mathbb E_{p_0(\bm x_0), U(\bm t; 0,1)} \lVert v_{\bm t}^\theta(\bm x_{\bm t}) - \dot {\bm x}_{\bm t} \rVert^2, \quad \text{where}\quad \bm x_{\bm t} = \bm t \phi_1^{\theta_m}(\bm x_0) + (1-\bm t) \bm x_0.
\end{equation}
In theory, each time the problem is solved, the paths generated by $v^{\theta_m}$ become more straight and thus need fewer discretization steps. However, since the model is trained on the outputs of another model, approximation errors can be introduced. As for direct distillation, one has to solve the flow matching ODE for each training sample. %

\section{Experiments}\label{sec:experiments}

\begin{figure}
    \centering
    \resizebox{!}{0.26\linewidth}{
    \begin{tikzpicture}
        \node at (0,0) {\includegraphics[width=70px]{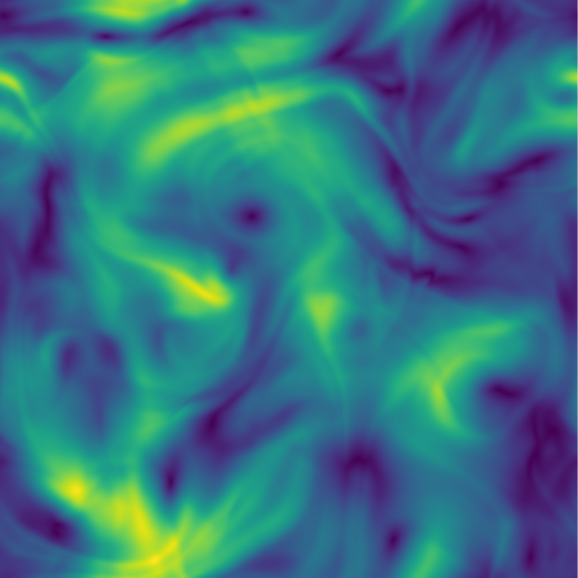}};
        \node at (1,-1) {\includegraphics[width=40px]{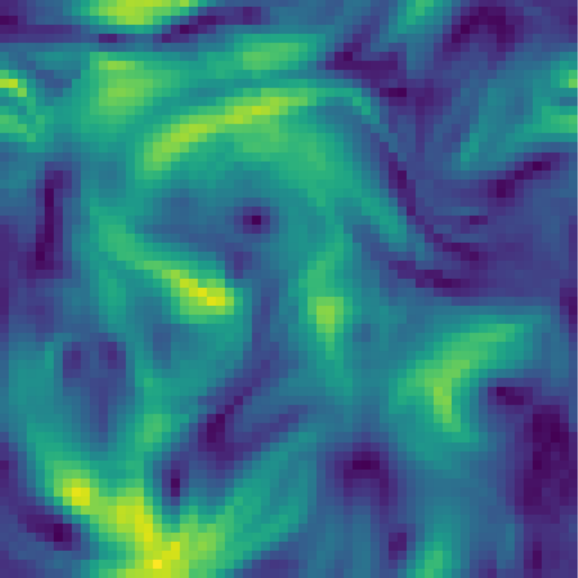}};
        \node at (0,-2.5) {\begin{tabular}{c} Downsampled NS (\textbf{dNSE}) \\ (showing magnitude of momentum) \end{tabular}};
        
        \node at (4, 0) {\includegraphics[width=80px]{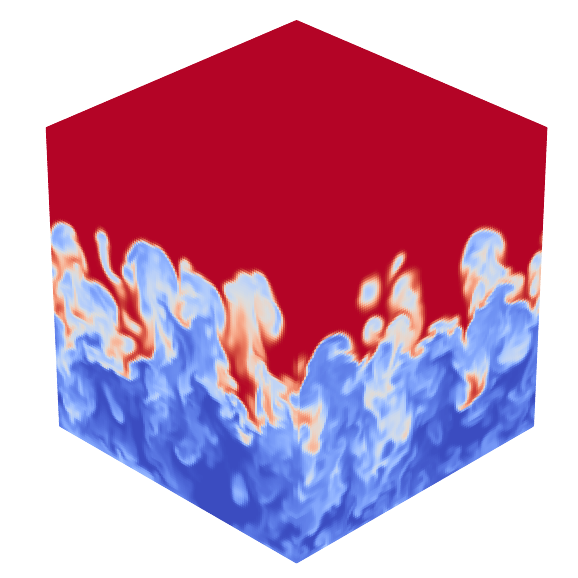}};
        \node at (5.3,-1) {\includegraphics[width=55px]{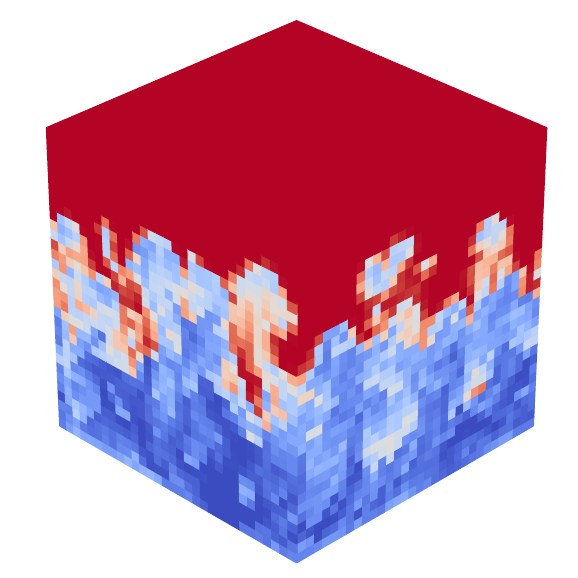}};
        \node at (6.5,-2.5) {\begin{tabular}{c} Downsampled/sliced RTI (\textbf{dRTI}/\textbf{sRTI}) \\ (showing density) \end{tabular}};
        
        \node at (8, 0) {\includegraphics[width=80px]{figures/rti3d_sample_original.pdf}};
        \draw[line width = 1mm] (9.24, -0.7) -- (8,-1.35) -- (8,0.23) 
                             -- (9.25,0.8) -- (9.24, -0.7); %
        \node at (9,-1) {\includegraphics[width=40px]{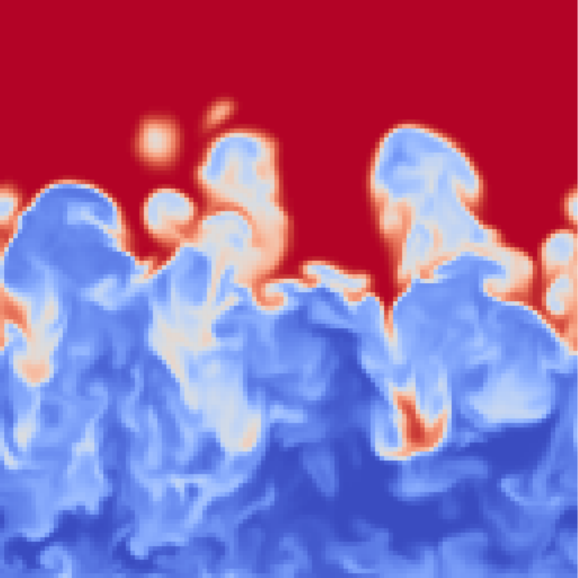}};
    \end{tikzpicture}}
    \caption{Samples and partial observations from the three datasets we are working with.}
    \label{fig:dataset-samples}
\end{figure}

Depending on the system, the transition distribution \eqref{eq:markov-chain} has lower or higher variance. %
In the low variance regime, we conduct experiments on downsampled NS (referred to as \textbf{dNSE}) and RTI (referred to as \textbf{dRTI}) simulations (see left and center plots in \Cref{fig:dataset-samples}). Our main goal is to show that probabilistic models can produce accurate forecasts while being as efficient as deterministic surrogate models. Additionally, we argue and demonstrate numerically that they perform better in the high-frequency energy spectrum and tend to produce sharper features.
On the other hand, if the measurements only partially capture the system dynamics, deterministic surrogate models become less and less reliable. In this regime, we train models to forecast two-dimensional ``slices'' of the three-dimensional RTI (referred to as \textbf{sRTI}, right plot in \Cref{fig:dataset-samples}). This scenario can occur, for instance, when generating inflow conditions for fluid flow simulations.
For both \textbf{dRTI} and \textbf{sRTI} datasets, we augment the observations by an additional channel containing a positional encoding for the vertical axis, and another channel containing the current time step of the simulation. The former is done because of the qualitatively different features at the bottom/top of a state, while the latter ensures that the model generates an appropriate amount of mixing for the given simulation time.

The NS dataset is taken from the ``PDEBench'' dataset (\cite{takamotoPDEBenchExtensiveBenchmark2022}, where we consider the low-viscosity case with shear and bulk viscosity $\eta = 10^{-8},~\zeta = 10^{-8}$, respectively). The computational domain $[0,1]^2$ with periodic boundary conditions is discretized by an equidistant $512^2$ grid, and for our purposes downsampled to $64^2$. The RTI dataset is taken from ``The Well'' \citep{ohanaWellLargeScaleCollection2024,cabotReynoldsNumberEffects2006}, with an Atwood number of $\mathrm{At} = 3/4$, defined on the domain $[0,1]^3$ and discretized by a $128^3$ grid. After downsampling (not done for the slices experiment), this becomes $32^3$. The domain is equipped with periodic boundary conditions in the horizontal directions, and free slip boundary conditions at the bottom and top. The RTI comes with 9 training and 2 test trajectories, each of length 119, and our models predict 5 steps into the future.

\begin{table}[]
    \centering
    \resizebox{0.9\linewidth}{!}{%
    \begin{NiceTabular}{rllcl}
        & \Block{1-2}{Training} & & & Inference \\
        \cmidrule{2-3}\cmidrule{5-5}
        Method & $\frac{\text{\#Iters}}{\text{second}}$ & FM ODE solved with & & \#Model evaluations \\
        \midrule
        Flow matching & 71.03 & --- & & 10 (\textbf{dNSE}, \textbf{dRTI}) / 40 (\textbf{sRTI}) \\ 
        Deterministic model & 69.11 & --- & & 1 \\
        Progressive distillation & 45.16 & --- & & 2 \\
        WGAN ($\frac{\text{\#}D \text{ iters}}{\text{\#}G \text{ iters}} = 5$) & 12.95 & --- & & 1 \\
        Rectified flow & 12.61 & 10 midpoint steps & & 4 \\
        Direct distillation & 12.56 & 10 midpoint steps & & 1 \\
        ADD ($\lambda \in (0, 1) $, $\frac{\text{\#}D \text{ iters}}{\text{\#}G \text{ iters}} = 5$) & 10.01 & 10 Euler steps & & 1 \\
        \bottomrule
    \end{NiceTabular}}
    \caption{Training iterations per seconds on the \textbf{dNSE} dataset with a batch size of 8 \& number of function evaluations used to generate all plots in this paper (same model architecture overall).}
    \label{tab:training-speed}
\end{table}

\subsection{Training}

We have trained flow matching models based on a standard UNet architecture; details are in the appendix. All experiments were conducted on a single A100 GPU. 

\paragraph{Training speed.}
The number of training iterations per second is recorded in \Cref{tab:training-speed}.
While flow matching training is efficient, extensions that require solutions of the flow matching ODE spend more time per iteration. This includes direct distillation, rectified flows and ADD.
However, for ADD, solving the ODE is only necessary for the generator. Typically, the generator is updated only every 5 or 10 iterations, which means that not much overhead is added by solving the ODE. Instead, the relatively low iteration speed stems from the gradient penalty that is computed for every discriminator update.
For progressive distillation, we note that while a single training iteration can be computed more efficiently than direct distillation, the ``exponential efficiency'' of this approach (i.e., training a model to make $N$ Euler steps requires only $\log_2 N$ training runs) is achieved only after multiple training runs: Most flow matching ODEs are solved to sufficient accuracy in 8-16 Euler steps, which would require at least 3-4 full training runs. 

\paragraph{Setup complexity.}
Direct distillation is the easiest to setup. It works well for problems that have clear successor states (e.g., \textbf{dNSE}), but for more nondeterministic transitions, it produces blurry features. The setup for progressive distillation is more involved. It should be noted, however, that a separate model is trained for each stage, allowing us to flexibly choose the desired accuracy.
Rectified flows are comparable with progressive distillation setup-wise due to the multi-stage setting.
Finally, ADD requires careful tuning of generator/discriminator learning rates, the gradient penalty and the ratio of generator to discriminator updates. 
However, tackling these issues can be rewarding, as it makes both stable predictions and generates physically-accurate features.

\begin{figure}
    \centering
    \includegraphics[width=\linewidth]{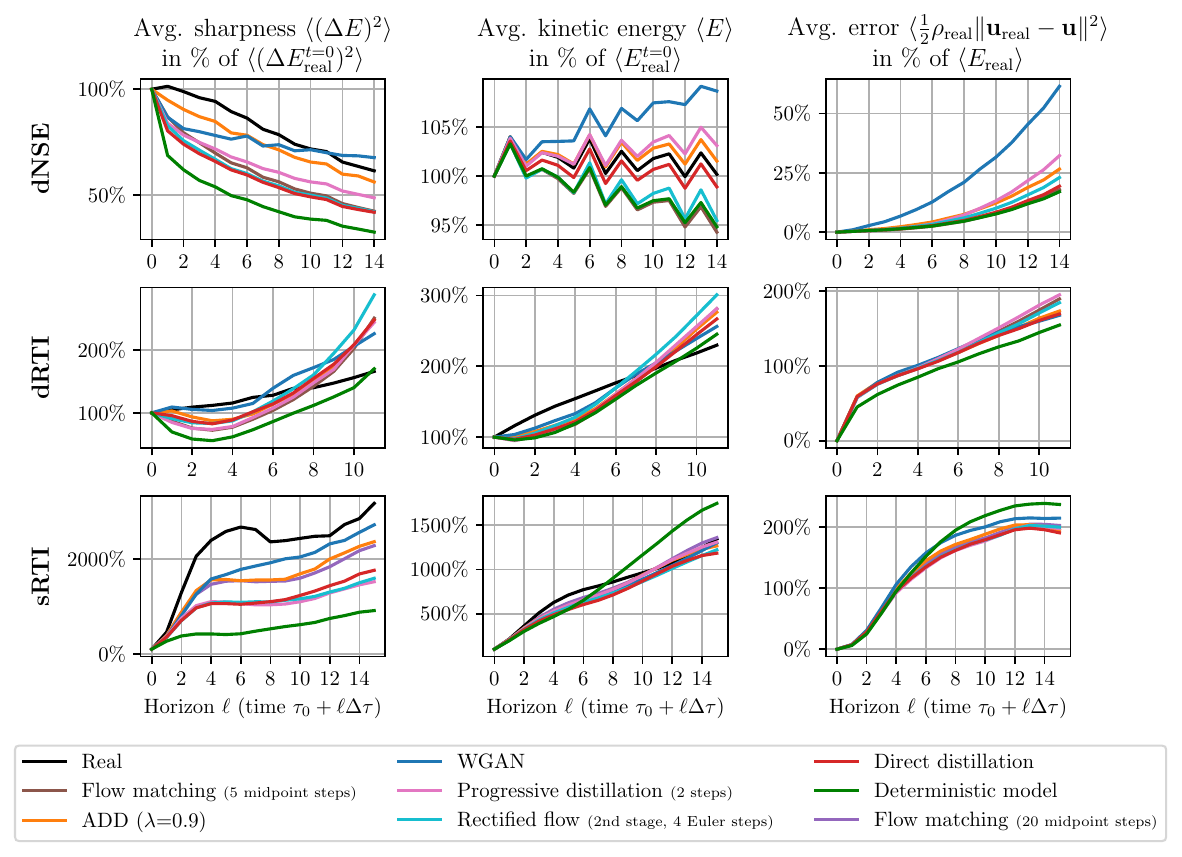}
    \caption{Statistics (sharpness, energy and error to real trajectory) of model predictions over prediction horizon. Results averaged over all initial conditions.}
    \label{fig:statistics}
\end{figure}

\subsection{Evaluation}

In order to provide a qualitative assessment of the trained models, we have shown their outputs after a single prediction step in \Cref{fig:single-prediction}, see also \Cref{fig:showcase-rti} for some sample trajectories on the \textbf{sRTI} dataset (we show more trajectories in the appendix, \Cref{fig:multiple-trajectories-ns,fig:multiple-trajectories-rti-full,fig:multiple-trajectories-rti-sliced}).

We first compare the deviation from the baseline trajectory over multiple steps in terms of averaged physical quantities composed of the density $\rho$ and velocity $\mathbf u$.
One way to define such a distance is to aggregate the pointwise error between the velocities, weighted by the real density:
\begin{equation}
    \langle \textstyle \frac 1 2  \rho_{\text{real}} \lVert \mathbf u_{\text{real}} - \mathbf u \rVert^2 \rangle,
    \label{eq:kinetic-energy-error}
\end{equation}
where $\langle \cdot \rangle$ denotes the average over the computational domain. In particular, we can relate this error to the kinetic energy, which is defined as $E = \langle \textstyle \frac 1 2 \rho \lVert \mathbf u \rVert^2 \rangle$.
The error \eqref{eq:kinetic-energy-error} is easy to compute, but it has its caveats. It is less sensitive to small-scale errors such as blurring, but sensitive to more qualitative changes, e.g., when shifting one of the states in space. For this, it is only really expressive in the case where the transition distribution has low variance. However, it allows us to quantify strong deviations from the baseline trajectory. We have plotted this deviation in the right column of \Cref{fig:statistics}. Here, we can observe that in particular the WGAN (ADD with $\lambda = 0$) diverges strongly for the \textbf{dNSE} dataset.

\begin{figure}
    \centering
    \includegraphics[width=\linewidth]{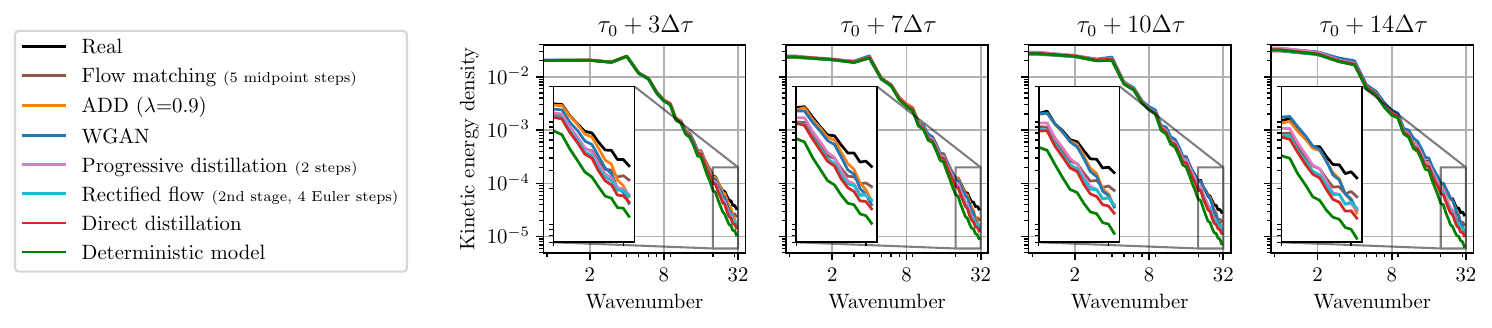}
    \caption{Spectra of the kinetic energy density on the \textbf{dNSE} dataset after multiple autoregressive prediction steps with equal initial conditions. The spectra are then averaged over all initial conditions. (See also \Cref{fig:kinetic-energy-density-ns-offset} in the appendix for further comparison.)}
    \label{fig:kinetic-energy-density-ns}
\end{figure}

It is apparent that the deterministic model produces blurry outputs, especially for sliced RTI, something that is expected given the arguments in \Cref{sec:deterministic-models}. 
We first study the energy spectrum focusing on the higher frequencies, representing the kinetic energy contained in smaller eddies. In \Cref{fig:kinetic-energy-density-ns} one can observe the drop in kinetic energy of the deterministic model for large wavenumbers on the \textbf{dNSE} dataset compared with the other models. As can be seen in the figure, the spread between the deterministic and the remaining forecasts even tends to increase over time.

Another measure of ``sharpness'' used in older autofocus applications~\citep{groenComparisonDifferentFocus1985} is given by the squared Laplacian of the kinetic energy,
quantifying how strongly values deviate from the average given by their neighbors. We have plotted this quantity in the left column of \Cref{fig:statistics}. The takeaway is similar to that of \Cref{fig:kinetic-energy-density-ns}: The deterministic model ranks lowest, while the WGAN produces the sharpest results. Interestingly, most methods tend to have lower sharpness than the real trajectory on each but the \textbf{dRTI} dataset (cf. \Cref{fig:multiple-trajectories-rti-full} in the appendix).

For a final point of reference, we compare the average kinetic energy in the center column of \Cref{fig:statistics}. For both \textbf{dRTI} and \textbf{sRTI} one can observe a strong increase over time, which is due to the fact that the potential energy of the initial state is transformed into kinetic energy as the fluids start mixing. This is captured by all models. For the \textbf{dNSE} dataset, a direct comparison with the kinetic energy of the baseline trajectory is more interesting: Some models (WGAN, progressive distillation, ADD with $\lambda = 0.9$) tend to introduce, while some others (FM, direct distillation, rectified FM, deterministic model) tend to remove energy from the system. %

\section{Conclusion}

We have discussed extensions to the flow matching paradigm for the probabilistic surrogate modeling of partially-observed dynamical systems. In particular, we have demonstrated and compared these methods on a set of challenging datasets, including simulations of the compressible Navier--Stokes equations and the Rayleigh--Taylor instability. We have found that while harder to tune, adversarial diffusion distillation provides the best results, especially with regards to fast sampling and the generation of plausible features. In contrast, if 2--4  more function evaluations are acceptable, progressive distillation is an attractive alternative that both has a simple setup and can be trained efficiently.

\acks{
HH acknowledges support by ``SAIL: SustAInable Lifecycle of Intelligent Socio-Technical Systems'' (Grant ID NW21-059D), funded by the Ministry of Culture and Science of the State of Northrhine Westphalia (NRW), Germany. SP acknowledges support by the European Union via the ERC Starting Grant
``KoOpeRaDE'' (Grant ID 101161457).
}

\bibliography{bibliography}

\begin{thebibliography}{43}
\providecommand{\natexlab}[1]{#1}
\providecommand{\url}[1]{\texttt{#1}}
\expandafter\ifx\csname urlstyle\endcsname\relax
  \providecommand{\doi}[1]{doi: #1}\else
  \providecommand{\doi}{doi: \begingroup \urlstyle{rm}\Url}\fi

\bibitem[Cabot and Cook(2006)]{cabotReynoldsNumberEffects2006}
William~H. Cabot and Andrew~W. Cook.
\newblock Reynolds number effects on {{Rayleigh}}--{{Taylor}} instability with possible implications for type {{Ia}} supernovae.
\newblock \emph{Nature Physics}, 2\penalty0 (8):\penalty0 562--568, August 2006.
\newblock ISSN 1745-2473, 1745-2481.

\bibitem[Cachay et~al.(2023)Cachay, Zhao, Joren, and Yu]{cachayDYffusionDynamicsinformedDiffusion2023}
Salva~R{\"u}hling Cachay, Bo~Zhao, Hailey Joren, and Rose Yu.
\newblock {{DYffusion}}: {{A Dynamics-informed Diffusion Model}} for {{Spatiotemporal Forecasting}}.
\newblock In \emph{Proceedings of the 37th {{Annual Conference}} on {{Neural Information Processing Systems}}}, 2023.

\bibitem[Chattopadhyay et~al.(2024)Chattopadhyay, Gray, Wu, Lowe, and He]{chattopadhyayOceanNetPrincipledNeural2024}
Ashesh Chattopadhyay, Michael Gray, Tianning Wu, Anna~B. Lowe, and Ruoying He.
\newblock {{OceanNet}}: A principled neural operator-based digital twin for regional oceans.
\newblock \emph{Scientific Reports}, 14\penalty0 (1):\penalty0 21181, September 2024.
\newblock ISSN 2045-2322.

\bibitem[Dhariwal and Nichol(2021)]{dhariwalDiffusionModelsBeat2021}
Prafulla Dhariwal and Alex Nichol.
\newblock Diffusion {{Models Beat GANs}} on {{Image Synthesis}}.
\newblock In \emph{Proceedings of the 35th {{Annual Conference}} on {{Neural Information Processing Systems}}}, volume~34, pages 8780--8794, June 2021.

\bibitem[{El-Gazzar} and van Gerven(2025)]{el-gazzarProbabilisticForecastingAutoregressive2025}
Ahmed {El-Gazzar} and Marcel van Gerven.
\newblock Probabilistic {{Forecasting}} via {{Autoregressive Flow Matching}}, March 2025.
\newblock Preprint.

\bibitem[Fromme et~al.(2025)Fromme, Harder, {Allen-Blanchette}, and Peitz]{frommeSurrogateModeling3D2025}
Fynn Fromme, Hans Harder, Christine {Allen-Blanchette}, and Sebastian Peitz.
\newblock Surrogate {{Modeling}} of {{3D Rayleigh-Benard Convection}} with {{Equivariant Autoencoders}}, September 2025.
\newblock Preprint.

\bibitem[Gao et~al.(2024)Gao, Kaltenbach, and Koumoutsakos]{gaoGenerativeLearningForecasting2024}
Han Gao, Sebastian Kaltenbach, and Petros Koumoutsakos.
\newblock Generative {{Learning}} for {{Forecasting}} the {{Dynamics}} of {{Complex Systems}}, February 2024.
\newblock Preprint.

\bibitem[Groen et~al.(1985)Groen, Young, and Ligthart]{groenComparisonDifferentFocus1985}
Frans C.~A. Groen, Ian~T. Young, and Guido Ligthart.
\newblock A comparison of different focus functions for use in autofocus algorithms.
\newblock \emph{Cytometry}, 6\penalty0 (2):\penalty0 81--91, March 1985.
\newblock ISSN 0196-4763, 1097-0320.

\bibitem[Guastoni and Vinuesa(2025)]{Guastoni2025}
Luca Guastoni and Ricardo Vinuesa.
\newblock A new perspective on the simulation of stochastic problems in fluid mechanics with diffusion models.
\newblock \emph{Nature Machine Intelligence}, 7\penalty0 (6):\penalty0 816--817, Jun 2025.
\newblock ISSN 2522-5839.

\bibitem[Ho et~al.(2020)Ho, Jain, and Abbeel]{hoDenoisingDiffusionProbabilistic2020}
Jonathan Ho, Ajay Jain, and Pieter Abbeel.
\newblock Denoising {{Diffusion Probabilistic Models}}.
\newblock In \emph{Proceedings of the 34th {{Annual Conference}} on {{Neural Information Processing Systems}}}, volume~33, December 2020.

\bibitem[Holderrieth et~al.(2025)Holderrieth, Havasi, Yim, Shaul, Gat, Jaakkola, Karrer, Chen, and Lipman]{holderrieth2025generatormatchinggenerativemodeling}
Peter Holderrieth, Marton Havasi, Jason Yim, Neta Shaul, Itai Gat, Tommi Jaakkola, Brian Karrer, Ricky T.~Q. Chen, and Yaron Lipman.
\newblock Generator matching: Generative modeling with arbitrary markov processes.
\newblock In \emph{The Thirteenth International Conference on Learning Representations}, 2025.

\bibitem[Kohl et~al.(2024)Kohl, Chen, and Thuerey]{kohlBenchmarkingAutoregressiveConditional2024}
Georg Kohl, Li-Wei Chen, and Nils Thuerey.
\newblock Benchmarking {{Autoregressive Conditional Diffusion Models}} for {{Turbulent Flow Simulation}}, December 2024.
\newblock Preprint.

\bibitem[Lam et~al.(2023)Lam, {Sanchez-Gonzalez}, Willson, Wirnsberger, Fortunato, Alet, Ravuri, Ewalds, {Eaton-Rosen}, Hu, Merose, Hoyer, Holland, Vinyals, Stott, Pritzel, Mohamed, and Battaglia]{lamLearningSkillfulMediumrange2023}
Remi Lam, Alvaro {Sanchez-Gonzalez}, Matthew Willson, Peter Wirnsberger, Meire Fortunato, Ferran Alet, Suman Ravuri, Timo Ewalds, Zach {Eaton-Rosen}, Weihua Hu, Alexander Merose, Stephan Hoyer, George Holland, Oriol Vinyals, Jacklynn Stott, Alexander Pritzel, Shakir Mohamed, and Peter Battaglia.
\newblock Learning skillful medium-range global weather forecasting.
\newblock \emph{Science}, 382\penalty0 (6677):\penalty0 1416--1421, December 2023.
\newblock ISSN 0036-8075, 1095-9203.

\bibitem[Li et~al.(2025)Li, Zhou, and Farimani]{liGenerativeLatentNeural2025}
Zijie Li, Anthony Zhou, and Amir~Barati Farimani.
\newblock Generative {{Latent Neural PDE Solver}} using {{Flow Matching}}, March 2025.
\newblock Preprint.

\bibitem[Li et~al.(2020)Li, Kovachki, Azizzadenesheli, Liu, Bhattacharya, Stuart, and Anandkumar]{liNeuralOperatorGraph2020}
Zongyi Li, Nikola Kovachki, Kamyar Azizzadenesheli, Burigede Liu, Kaushik Bhattacharya, Andrew Stuart, and Anima Anandkumar.
\newblock Neural {{Operator}}: {{Graph Kernel Network}} for {{Partial Differential Equations}}.
\newblock In \emph{Proceedings of the {{ICLR}} 2020 {{Workshop}} on {{Integration}} of {{Deep Neural Models}} and {{Differential Equations}}}, March 2020.

\bibitem[Li et~al.(2021)Li, Kovachki, Azizzadenesheli, Liu, Bhattacharya, Stuart, and Anandkumar]{liFourierNeuralOperator2021a}
Zongyi Li, Nikola Kovachki, Kamyar Azizzadenesheli, Burigede Liu, Kaushik Bhattacharya, Andrew Stuart, and Anima Anandkumar.
\newblock Fourier {{Neural Operator}} for {{Parametric Partial Differential Equations}}.
\newblock In \emph{Proceedings of the 9th {{International Conference}} on {{Learning Representations}}}, May 2021.

\bibitem[Lim and Ye(2017)]{limGeometricGAN2017}
Jae~Hyun Lim and Jong~Chul Ye.
\newblock Geometric {{GAN}}, May 2017.
\newblock Preprint.

\bibitem[Lin et~al.(2024)Lin, Wang, and Yang]{linSDXLLightningProgressiveAdversarial2024}
Shanchuan Lin, Anran Wang, and Xiao Yang.
\newblock {{SDXL-Lightning}}: {{Progressive Adversarial Diffusion Distillation}}, March 2024.
\newblock Preprint.

\bibitem[Lipman et~al.(2023)Lipman, Chen, {Ben-Hamu}, Nickel, and Le]{lipmanFlowMatchingGenerative2023}
Yaron Lipman, Ricky T.~Q. Chen, Heli {Ben-Hamu}, Maximilian Nickel, and Matt Le.
\newblock Flow {{Matching}} for {{Generative Modeling}}.
\newblock In \emph{Proceedings of the 11th {{International Conference}} on {{Learning Representations}}}, 2023.

\bibitem[Lipman et~al.(2024)Lipman, Havasi, Holderrieth, Shaul, Le, Karrer, Chen, {Lopez-Paz}, {Ben-Hamu}, and Gat]{lipmanFlowMatchingGuide2024}
Yaron Lipman, Marton Havasi, Peter Holderrieth, Neta Shaul, Matt Le, Brian Karrer, Ricky T.~Q. Chen, David {Lopez-Paz}, Heli {Ben-Hamu}, and Itai Gat.
\newblock Flow {{Matching Guide}} and {{Code}}, December 2024.
\newblock Preprint.

\bibitem[Liu et~al.(2023)Liu, Gong, and Liu]{liuFlowStraightFast2023}
Xingchao Liu, Chengyue Gong, and Qiang Liu.
\newblock Flow {{Straight}} and {{Fast}}: {{Learning}} to {{Generate}} and {{Transfer Data}} with {{Rectified Flow}}.
\newblock In \emph{Proceedings of the 11th {{International Conference}} on {{Learning Representations}}}, 2023.

\bibitem[Liu et~al.(2024)Liu, Zhang, Ma, Peng, and Liu]{liuInstaFlowOneStep2024}
Xingchao Liu, Xiwen Zhang, Jianzhu Ma, Jian Peng, and Qiang Liu.
\newblock {{InstaFlow}}: {{One Step}} is {{Enough}} for {{High-Quality Diffusion-Based Text-to-Image Generation}}.
\newblock In \emph{Proceedings of the 12th {{International Conference}} on {{Learning Representations}}}, March 2024.

\bibitem[Luo et~al.(2024)Luo, Wu, Wang, Xie, Yue, and Tang]{luoDiffFluidPlainDiffusion2024}
Dongyu Luo, Jianyu Wu, Jing Wang, Hairun Xie, Xiangyu Yue, and Shixiang Tang.
\newblock {{DiffFluid}}: {{Plain Diffusion Models}} are {{Effective Predictors}} of {{Flow Dynamics}}, September 2024.
\newblock Preprint.

\bibitem[Nichol and Dhariwal(2021)]{nicholImprovedDenoisingDiffusion2021}
Alexander~Quinn Nichol and Prafulla Dhariwal.
\newblock Improved {{Denoising Diffusion Probabilistic Models}}.
\newblock In \emph{Proceedings of the 38th {{International Conference}} on {{Machine Learning}}}, pages 8162--8171. PMLR, July 2021.

\bibitem[Ohana et~al.(2024)Ohana, McCabe, Meyer, Morel, Agocs, Beneitez, Berger, Burkhart, Dalziel, Fielding, Fortunato, Goldberg, Hirashima, Jiang, Kerswell, Maddu, Miller, Mukhopadhyay, Nixon, Shen, Watteaux, Blancard, Rozet, Parker, Cranmer, and Ho]{ohanaWellLargeScaleCollection2024}
Ruben Ohana, Michael McCabe, Lucas Meyer, Rudy Morel, Fruzsina~J Agocs, Miguel Beneitez, Marsha Berger, Blakesley Burkhart, Stuart~B Dalziel, Drummond~B Fielding, Daniel Fortunato, Jared~A Goldberg, Keiya Hirashima, Yan-Fei Jiang, Rich~R Kerswell, Suryanarayana Maddu, Jonah Miller, Payel Mukhopadhyay, Stefan~S Nixon, Jeff Shen, Romain Watteaux, Bruno R{\'e}galdo-Saint Blancard, Fran{\c c}ois Rozet, Liam~H Parker, Miles Cranmer, and Shirley Ho.
\newblock The {{Well}}: A {{Large-Scale Collection}} of {{Diverse Physics Simulations}} for {{Machine Learning}}.
\newblock In \emph{Proceedings of the 38th {{Annual Conference}} on {{Neural Information Processing Systems}}}, volume~37, pages 44989--45037, 2024.

\bibitem[Oommen et~al.(2025)Oommen, Khodakarami, Bora, Wang, and Karniadakis]{oommenLearningTurbulentFlows2025}
Vivek Oommen, Siavash Khodakarami, Aniruddha Bora, Zhicheng Wang, and George~Em Karniadakis.
\newblock Learning {{Turbulent Flows}} with {{Generative Models}}: {{Super-resolution}}, {{Forecasting}}, and {{Sparse Flow Reconstruction}}, September 2025.
\newblock Preprint.

\bibitem[Price et~al.(2025)Price, {Sanchez-Gonzalez}, Alet, Andersson, {El-Kadi}, Masters, Ewalds, Stott, Mohamed, Battaglia, Lam, and Willson]{priceProbabilisticWeatherForecasting2025}
Ilan Price, Alvaro {Sanchez-Gonzalez}, Ferran Alet, Tom~R. Andersson, Andrew {El-Kadi}, Dominic Masters, Timo Ewalds, Jacklynn Stott, Shakir Mohamed, Peter Battaglia, Remi Lam, and Matthew Willson.
\newblock Probabilistic weather forecasting with machine learning.
\newblock \emph{Nature}, 637\penalty0 (8044):\penalty0 84--90, January 2025.
\newblock ISSN 0028-0836, 1476-4687.

\bibitem[Salimans and Ho(2022)]{salimansProgressiveDistillationFast2022}
Tim Salimans and Jonathan Ho.
\newblock Progressive {{Distillation}} for {{Fast Sampling}} of {{Diffusion Models}}.
\newblock In \emph{Proceedings of the 10th {{International Conference}} on {{Learning Representations}}}, 2022.

\bibitem[Sauer et~al.(2023)Sauer, Lorenz, Blattmann, and Rombach]{sauerAdversarialDiffusionDistillation2023}
Axel Sauer, Dominik Lorenz, Andreas Blattmann, and Robin Rombach.
\newblock Adversarial {{Diffusion Distillation}}.
\newblock In \emph{Proceedings of the 18th {{European Conference}} on {{Computer Vision}}}, November 2023.

\bibitem[Shysheya et~al.(2024)Shysheya, Diaconu, and Bergamin]{shysheyaConditionalDiffusionModels2024}
Aliaksandra Shysheya, Cristiana Diaconu, and Federico Bergamin.
\newblock On conditional diffusion models for {{PDE}} simulations.
\newblock In \emph{Proceedings of the 38th {{Annual Conference}} on {{Neural Information Processing Systems}}}, volume~38, 2024.

\bibitem[{Sohl-Dickstein} et~al.(2015){Sohl-Dickstein}, Weiss, Maheswaranathan, and Ganguli]{sohl-dicksteinDeepUnsupervisedLearning2015}
Jascha {Sohl-Dickstein}, Eric~A. Weiss, Niru Maheswaranathan, and Surya Ganguli.
\newblock Deep {{Unsupervised Learning}} using {{Nonequilibrium Thermodynamics}}.
\newblock In \emph{Proceedings of the 32nd {{International Conference}} on {{Machine Learning}}}, volume~37, pages 2256--2265. JMLR.org, November 2015.

\bibitem[Song et~al.(2021)Song, Meng, and Ermon]{songDenoisingDiffusionImplicit2021}
Jiaming Song, Chenlin Meng, and Stefano Ermon.
\newblock Denoising {{Diffusion Implicit Models}}.
\newblock In \emph{Proceedings of the 9th {{International Conference}} on {{Learning Representations}}}, 2021.

\bibitem[Song and Dhariwal(2024)]{songImprovedTechniquesTraining2024}
Yang Song and Prafulla Dhariwal.
\newblock Improved {{Techniques}} for {{Training Consistency Models}}.
\newblock In \emph{Proceedings of the 12th {{International Conference}} on {{Learning Representations}} ({{ICLR}})}, volume~12, 2024.

\bibitem[Song and Ermon(2019)]{songGenerativeModelingEstimating2019}
Yang Song and Stefano Ermon.
\newblock Generative {{Modeling}} by {{Estimating Gradients}} of the {{Data Distribution}}.
\newblock In \emph{Proceedings of the 33rd {{Annual Conference}} on {{Neural Information Processing Systems}}}, volume~32, pages 11895--11907. Curran Associates, Inc., 2019.

\bibitem[Song et~al.(2023)Song, Dhariwal, Chen, and Sutskever]{songConsistencyModels2023}
Yang Song, Prafulla Dhariwal, Mark Chen, and Ilya Sutskever.
\newblock Consistency {{Models}}.
\newblock In \emph{Proceedings of the 40th {{International Conference}} on {{Machine Learning}}}, volume 202 of \emph{Proceedings of {{Machine Learning Research}}}, May 2023.

\bibitem[Srinivasan et~al.(2019)Srinivasan, Guastoni, Azizpour, Schlatter, and Vinuesa]{PhysRevFluids.4.054603}
P.~A. Srinivasan, L.~Guastoni, H.~Azizpour, P.~Schlatter, and R.~Vinuesa.
\newblock Predictions of turbulent shear flows using deep neural networks.
\newblock \emph{Phys. Rev. Fluids}, 4:\penalty0 054603, May 2019.

\bibitem[Takamoto et~al.(2022)Takamoto, Praditia, Leiteritz, MacKinlay, Alesiani, Pfl{\"u}ger, and Niepert]{takamotoPDEBenchExtensiveBenchmark2022}
Makoto Takamoto, Timothy Praditia, Raphael Leiteritz, Dan MacKinlay, Francesco Alesiani, Dirk Pfl{\"u}ger, and Mathias Niepert.
\newblock {{PDEBench}}: {{An Extensive Benchmark}} for {{Scientific Machine Learning}}.
\newblock In \emph{Proceedings of the 36th {{Annual Conference}} on {{Neural Information Processing Systems}}}, volume~35, pages 1596--1611, 2022.

\bibitem[Takens(1980)]{takensDetecting1980}
Floris Takens.
\newblock Detecting strange attractors in turbulence.
\newblock In \emph{Dynamical Systems and Turbulence, Warwick 1980: proceedings of a symposium held at the University of Warwick 1979/80}, pages 366--381. Springer, 1980.

\bibitem[Vishwasrao et~al.(2025)Vishwasrao, Gutha, Cremades, Wijk, Patil, Gorle, McKeon, Azizpour, and Vinuesa]{vishwasraoDiffSPORTDiffusionbasedSensor2025}
Abhijeet Vishwasrao, Sai Bharath~Chandra Gutha, Andres Cremades, Klas Wijk, Aakash Patil, Catherine Gorle, Beverley~J. McKeon, Hossein Azizpour, and Ricardo Vinuesa.
\newblock Diff-{{SPORT}}: {{Diffusion-based Sensor Placement Optimization}} and {{Reconstruction}} of {{Turbulent}} flows in urban environments, May 2025.
\newblock Preprint.

\bibitem[Vlachas et~al.(2018)Vlachas, Byeon, Wan, Sapsis, and Koumoutsakos]{vlachasDatadrivenForecastingHighdimensional2018}
Pantelis~R. Vlachas, Wonmin Byeon, Zhong~Y. Wan, Themistoklis~P. Sapsis, and Petros Koumoutsakos.
\newblock Data-driven forecasting of high-dimensional chaotic systems with long short-term memory networks.
\newblock \emph{Proceedings of the Royal Society A: Mathematical, Physical and Engineering Sciences}, 474\penalty0 (2213):\penalty0 20170844, May 2018.
\newblock ISSN 1364-5021, 1471-2946.

\bibitem[{Watt-Meyer} et~al.(2023){Watt-Meyer}, Dresdner, McGibbon, Clark, Henn, Duncan, Brenowitz, Kashinath, Pritchard, Bonev, Peters, and Bretherton]{watt-meyerACEFastSkillful2023}
Oliver {Watt-Meyer}, Gideon Dresdner, Jeremy McGibbon, Spencer~K. Clark, Brian Henn, James Duncan, Noah~D. Brenowitz, Karthik Kashinath, Michael~S. Pritchard, Boris Bonev, Matthew~E. Peters, and Christopher~S. Bretherton.
\newblock {{ACE}}: {{A}} fast, skillful learned global atmospheric model for climate prediction.
\newblock In \emph{Proceedings of the {{NeurIPS}} 2023 {{Workshop}} on {{Tackling Climate Change}} with {{Machine Learning}}}, December 2023.

\bibitem[Xiao et~al.(2022)Xiao, Kreis, and Vahdat]{xiaoTacklingGenerativeLearning2022}
Zhisheng Xiao, Karsten Kreis, and Arash Vahdat.
\newblock Tackling the {{Generative Learning Trilemma}} with {{Denoising Diffusion GANs}}.
\newblock In \emph{Proceedings of the 10th {{International Conference}} on {{Learning Representations}}}, 2022.

\bibitem[Yang and Sommer(2023)]{yangDenoisingDiffusionModel2023}
Gefan Yang and Stefan Sommer.
\newblock A {{Denoising Diffusion Model}} for {{Fluid Field Prediction}}, 2023.
\newblock Preprint.

\end{thebibliography}

\newpage

\appendix

\section{Additional content}

\begin{figure}[H]
    \centering
    \includegraphics[width=0.65\linewidth]{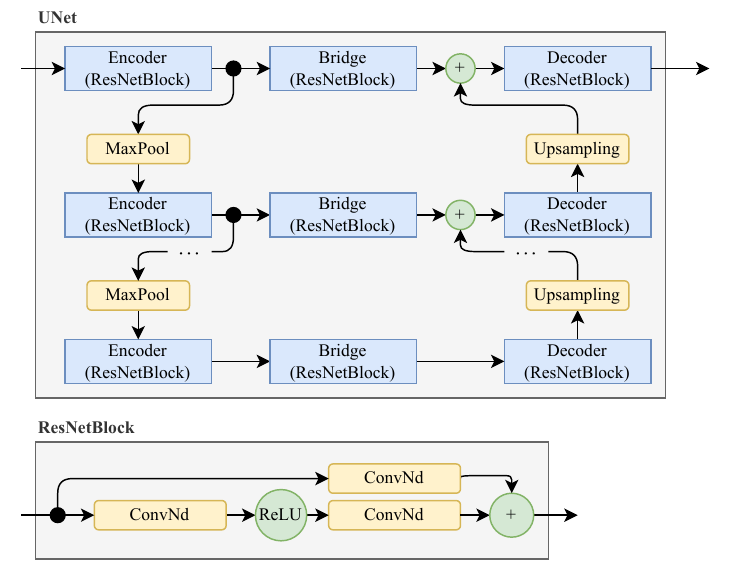}
    \caption{UNet architecture.}
    \label{fig:architecture}
\end{figure}

\begin{figure}[H]
    \centering
    \includegraphics[width=\linewidth]{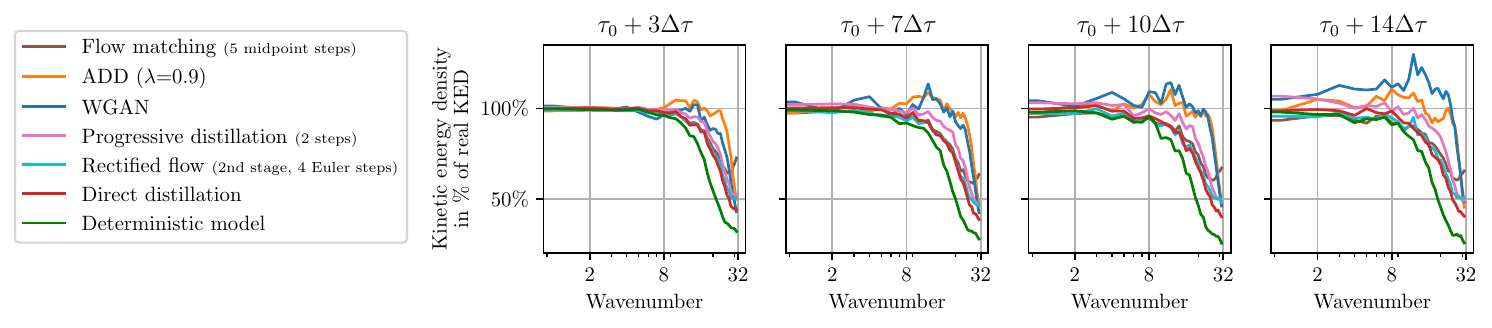}
    \caption{Kinetic energy densities from \Cref{fig:kinetic-energy-density-ns} in relationship to the real kinetic energy density.}
    \label{fig:kinetic-energy-density-ns-offset}
\end{figure}

\begin{figure}[H]
    \centering
    \includegraphics[width=\linewidth]{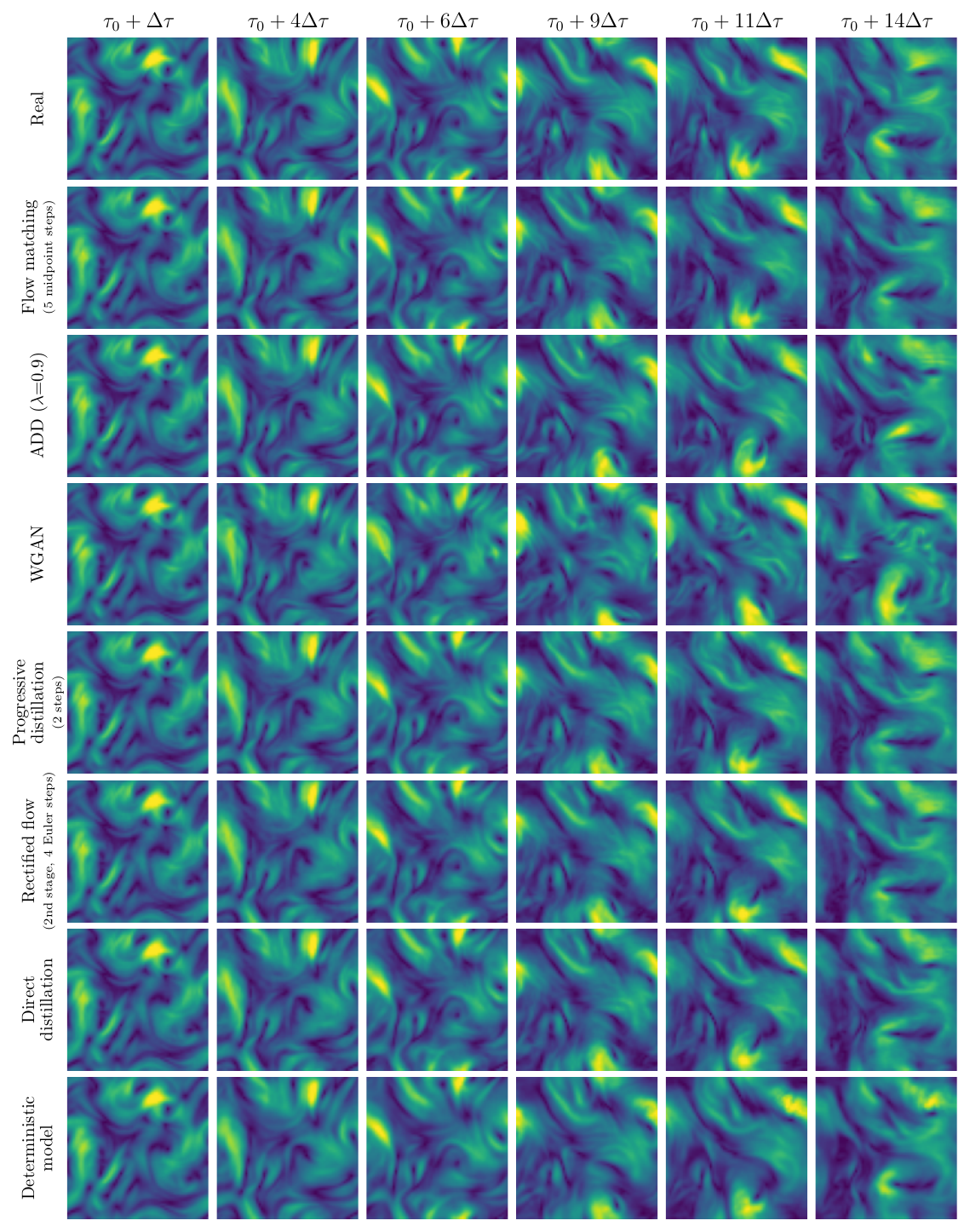}
    \caption{Multiple trajectories on the \textbf{dNSE} dataset.}
    \label{fig:multiple-trajectories-ns}
\end{figure}

\begin{figure}[H]
    \centering
    \includegraphics[width=\linewidth]{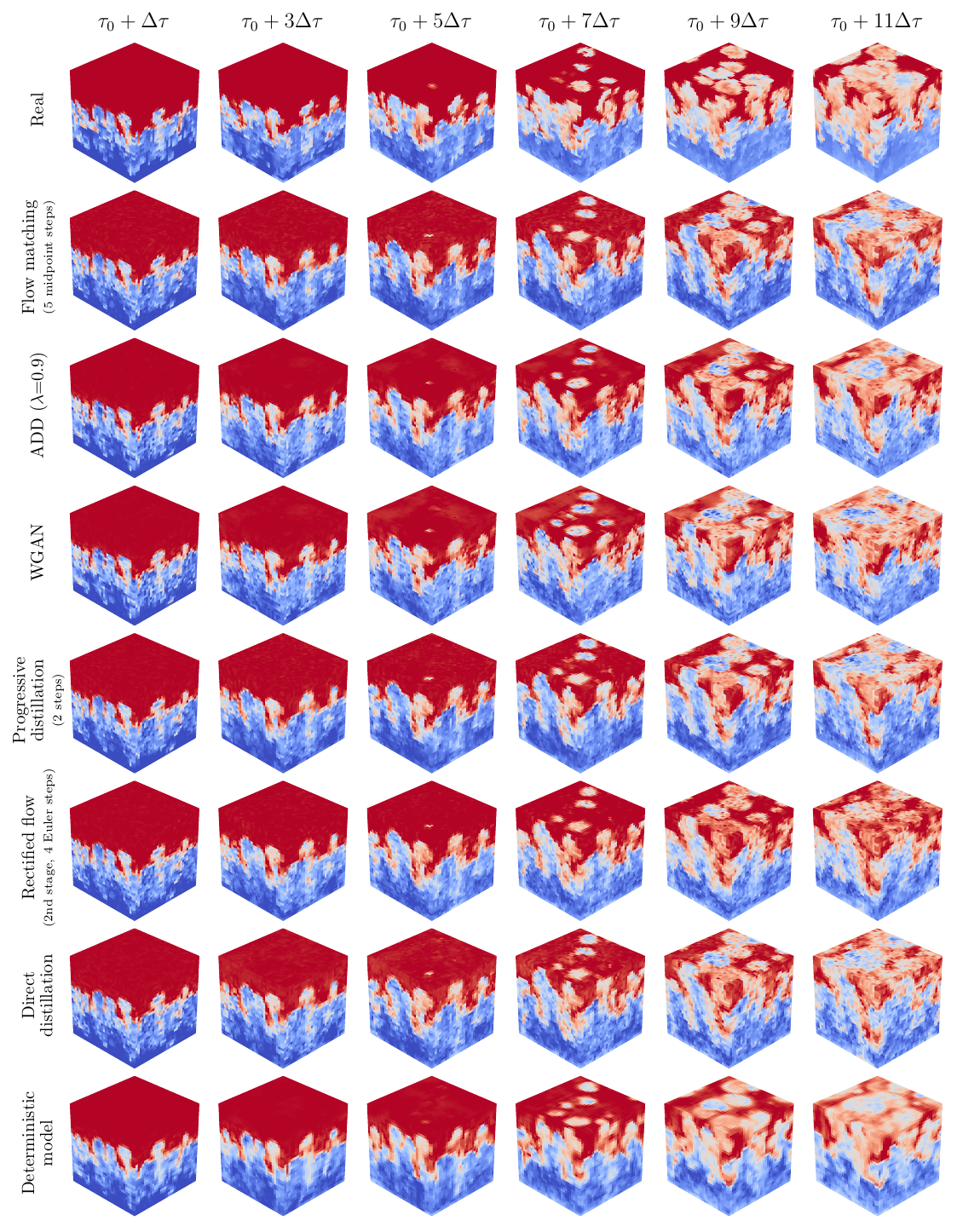}
    \caption{Multiple trajectories on the \textbf{dRTI} dataset.}
    \label{fig:multiple-trajectories-rti-full}
\end{figure}

\begin{figure}[H]
    \centering
    \includegraphics[width=\linewidth]{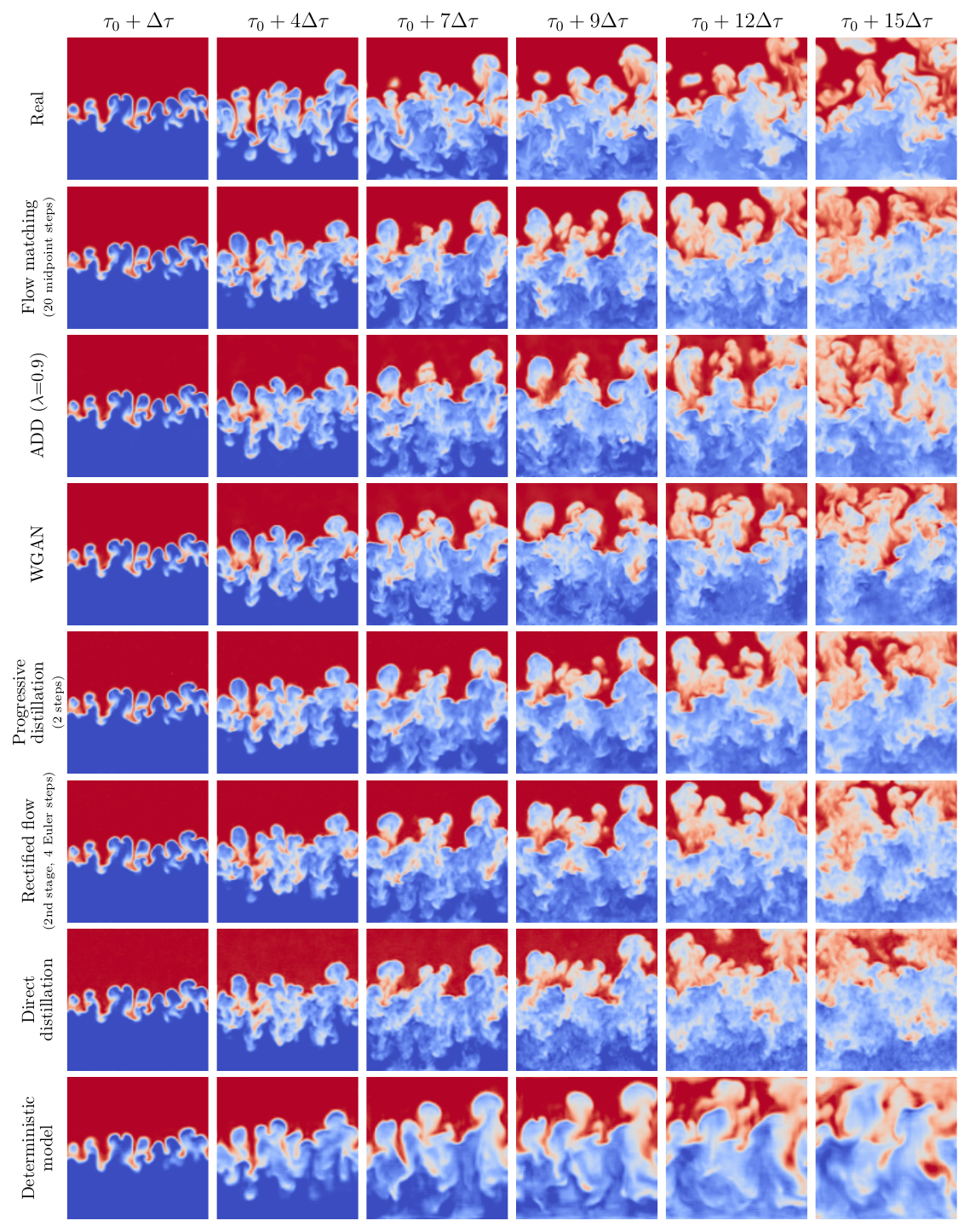}
    \caption{Multiple trajectories on the \textbf{sRTI} dataset.}
    \label{fig:multiple-trajectories-rti-sliced}
\end{figure}

Here we collect additional material that is helpful for understanding the main paper. In Remark \ref{rem:proof-of-equivalent-optimization-problems}, we argue why determinsitic models learn the expected value. The architecture of our UNet is shown in \Cref{fig:architecture}. In \Cref{fig:kinetic-energy-density-ns-offset}, we relate the kinetic energy density on the \textbf{dNSE} dataset to the reference. \Cref{tab:training-settings} contains details for our training setup. Finally, full trajectories are plotted in \Cref{fig:multiple-trajectories-ns,fig:multiple-trajectories-rti-full,fig:multiple-trajectories-rti-sliced}.

\begin{remark}
    The equivalence of \eqref{eq:deterministic-optimization-problem} and \eqref{eq:deterministic-optimization-problem-equivalent} can be shown as follows.
    Decompose \eqref{eq:deterministic-optimization-problem} as 
    \begin{align}
        \E_{p_1(\bm y_{k+1}, \bm y_k)} \lVert w^\theta(\bm y_k) \rVert^2 - 2 \E_{p_1(\bm y_{k+1}, \bm y_k)} [ w^\theta(\bm y_k) \cdot \bm y_{k+1} ] + \E_{p_1(\bm y_{k+1}, \bm y_k)}\lVert \bm y_{k+1} \rVert^2,
        \label{eq:deterministic-optimization-problem-equivalent-prf}
    \end{align}
    where the last term is independent of $\theta$. Moreover, we have
    \begin{align*}
        \E_{p_1(\bm y_{k+1}, \bm y_k)} \lVert w^\theta(\bm y_k) \rVert^2 &= \E_{p_1(\bm y_k)} \lVert w^\theta(\bm y_k)\rVert^2 \\
        \text{and}\quad \E_{p_1(\bm y_{k+1}, \bm y_k)} [ w^\theta(\bm y_k) \cdot \bm y_{k+1} ] &= \E_{p_1(\bm y_{k})} [ w^\theta(\bm y_k) \cdot \E_{p_1(\bm y_{k+1}|\bm y_k)} [\bm y_{k+1}] ]
    \end{align*}
    by independence of $w^\theta(\bm y_k)$ from $\bm y_{k+1}$.
    Substituting the right-hand sides into \eqref{eq:deterministic-optimization-problem-equivalent-prf} and changing the constant term to
    $\E_{p_1(\bm y_{k})} \lVert \E_{p_1(\bm y_{k+1}|\bm y_k)}[\bm y_{k+1}] \rVert^2,$
    we can rewrite the expression to yield \eqref{eq:deterministic-optimization-problem-equivalent}.
    \label{rem:proof-of-equivalent-optimization-problems}
\end{remark} 

\newpage

\begin{table}[H]
    \centering
    \resizebox{\textwidth}{!}{%
    \begin{NiceTabular}{ccccccccc}

        \toprule
         \Block{4-1}{Flow \\ Matching}
             & Dataset 
             & \#Iters 
             & $\frac{\text{\#Iters}}{\text{second}}$ 
             & Batch
             & \Block{1-2}{Learning rate} 
             & & Channels$^1$ 
             & \#Params\\
         \cmidrule{2-9}
            & dNSE & 820k & 27.85 & 32 & \Block{1-2}{1e-5} & & $128, 196, 196$ & 4,4m  \\
            & dRTI & 70k & 3.46 & 8 & \Block{1-2}{1e-5} & & $128, 196$ & 6,4m \\
            & sRTI & 350k & 7.11 & 32 & \Block{1-2}{1e-5} & & $128, 256, 256, 128$ & 8,6m \\
        \midrule
            
        \\[5pt]

        \cmidrule{1-7}
        \Block{4-1}{Deterministic \\ Model} 
           & Dataset 
           & \#Iters 
           & $\frac{\text{\#Iters}}{\text{second}}$ 
           & Batch
           & \Block{1-2}{Learning rate} & & \\
        \cmidrule{2-7}
           & dNSE  & 200k & 69.11 & 8 & \Block{1-2}{1e-5} \\
           & dRTI & 60k  & 10.33 & 8 & \Block{1-2}{1e-5} \\
           & sRTI & 500k & 25.54 & 8 & \Block{1-2}{1e-5} \\
        \cmidrule{1-7}

        \\[5pt]
        
        \cmidrule{1-7}
        \Block{4-1}{Direct \\ Distillation$^2$} 
            & Dataset & \#Iters 
            & $\frac{\text{\#Iters}}{\text{second}}$ 
            & Batch
            & \Block{1-2}{Learning rate} \\
        \cmidrule{2-7}
            & dNSE  & 100k & 12.56 & 8 & \Block{1-2}{1e-5} \\
            & dRTI &  40k & 3.28 & 4 & \Block{1-2}{1e-5} \\
            & sRTI & 350k & 6.07 & 4 & \Block{1-2}{1e-5} \\
        \cmidrule{1-7}

        \\[5pt]
        
        \cmidrule{1-8}
         \Block{4-1}{Progressive \\ Distillation}
             & Dataset 
             & $\frac{\text{\#Iters}}{\text{stage}}$ 
             & $\frac{\text{\#Iters}}{\text{second}}$ 
             & Batch
             & \Block{1-2}{Learning rate} 
             & & Stages \\
         \cmidrule{2-8}
             & dNSE & 100k & 45.16 & 8 & \Block{1-2}{1e-5} 
                & & $m=16, 8, 4, 2, 1$ \\
             & dRTI & 30k & 6.58 & 8 & \Block{1-2}{1e-5} 
                & & $m=16, 8, 4, 2, 1$ \\
             & sRTI & 100k & 14.71 & 8 & \Block{1-2}{1e-5} 
                & & $m=16, 8, 4, 2, 1$ \\
        \cmidrule{1-8}

        \\[5pt]
        
        \cmidrule{1-9}
         \Block{4-1}{Rectifiying \\ Flows$^2$ }
            & Dataset 
            & $\frac{\text{\#Iters}}{\text{stage}}$ 
            & $\frac{\text{\#Iters}}{\text{second}}$ 
            & Batch
            & \Block{1-2}{Learning rate} 
            & & \#Stages \\
         \cmidrule{2-9}
             & dNSE  & 100k & 12.61 & 8 & \Block{1-2}{1e-5} & & 2 \\
             & dRTI &  20k & 1.77  & 8 & \Block{1-2}{1e-5} & & 2 \\
             & sRTI & 100k & 3.73  & 8 & \Block{1-2}{1e-5} & & 2 \\
        \cmidrule{1-9}
        
        \\[5pt]

        \midrule
         \Block{7-1}{ADD$^3$} 
             & & & & & \Block{1-2}{Learning rate} \\ \cmidrule{6-7}
             & Dataset 
             & \#Iters 
             & $\frac{\text{\#Iters}}{\text{second}}$
             & Batch
             & $G$ & $D$ 
             & $\frac{\text{\#}D \text{ iters}}{\text{\#}G \text{ iters}}$
             & $\gamma$ \\ 
         \cmidrule{2-9}
             & \Block[t]{2-1}{dNSE} 
                & 150k & 10.01$^4$/12.95$^5$ & 8 & 5e-6 & 5e-5 & 5 & 5.0 \\
              & & +20k &                      &   & 1e-6 & 1e-5 &   &     \\
             & dRTI & 35k & 1.45$^4$/3.33$^5$ & 4 & 1e-6 & 1e-5 & 10 & 25.0 \\
             & \Block[t]{2-1}{sRTI} & 150k  & 3.16$^4$/3.45$^5$ & 8 & 1e-5 & 1e-4 & 5 & 25.0 \\
                                  & & +10k &                     &   & 1e-6 & 5e-5 &   &      \\
        \bottomrule
    \end{NiceTabular}}
    \caption{Training settings for each method and dataset. All models are based on the same UNet architecture given by the corresponding flow matching model. $^1$Channels along the downsampling path of the UNet. $^2$FM ODE solved with 10 midpoint steps. $^3$FM ODE solved with 10 Euler steps. $^4$With distillation loss. $^5$Without distillation loss, i.e., WGAN.}
    \label{tab:training-settings}
\end{table}

\end{document}